\documentclass{article}
\usepackage{ijcai25}
\usepackage{times}
\usepackage{soul}
\usepackage{url}
\usepackage[hidelinks]{hyperref}
\usepackage[utf8]{inputenc}
\usepackage[small]{caption}
\usepackage{amsmath}
\usepackage{amsthm}
\usepackage{booktabs}
\usepackage{algorithm}
\usepackage{algorithmic}
\usepackage[switch]{lineno}
\usepackage{graphicx}  
\usepackage{float}  
\usepackage{subfigure}  
\usepackage[normalem]{ulem}
\usepackage{flushend}
\usepackage{color}
\useunder{\uline}{\ul}{}

\newtheorem{example}{Example}

\title{RLBayes: a Bayesian Network Structure Learning Algorithm via \\ Reinforcement Learning-Based Search Strategy}

\author{
Mingcan Wang\and Junchang Xin \and Luxuan Qu\and Qi Chen \And Zhiqiong Wang
\\
\affiliations
School of Computer Science and Engineering, Northeastern University, China\\
Key Laboratory of Big Data Management and Analytics (Liaoning Province), Northeastern University, China\\
Ara Institute of Canterbury International Engineering College, Shenyang Jianzhu University, China\\
College of Medicine and Biological Information Engineering, Northeastern University, China\\
\emails
}

\begin{document}
\maketitle

\begin{abstract}
The score-based structure learning of Bayesian network (BN) is an effective way to learn BN models, which are regarded as some of the most compelling probabilistic graphical models in the field of representation and reasoning under uncertainty. However, the search space of structure learning grows super-exponentially as the number of variables increases, which makes BN structure learning an NP-hard problem, as well as a combination optimization problem (COP). Despite the successes of many heuristic methods on it, the results of the structure learning of BN are usually unsatisfactory. Inspired by $Q$-learning, in this paper, a Bayesian network structure learning algorithm via reinforcement learning-based (RL-based) search strategy is proposed, namely RLBayes. The method borrows the idea of RL and tends to record and guide the learning process by a dynamically maintained $Q$-table. By creating and maintaining the dynamic $Q$-table, RLBayes achieve storing the unlimited search space within limited space, thereby achieving the structure learning of BN via $Q$-learning. Not only is it theoretically proved that RLBayes can converge to the global optimal BN structure, but also it is experimentally proved that RLBayes has a better effect than almost all other heuristic search algorithms. Codes available at Appendix. 
\end{abstract}

\section{Introduction}\label{sec1}

Bayesian network (BN) is a probabilistic graph model which is widely used in engineering, bioinformatics and many other fields~\cite{BN0,BN1}. The structure learning of BN refers to determining the BN structure that best satisfies a given dataset. Accurate structure learning forges the supporting construction of the research of BN~\cite{BNSL,SWE2025,ICASSP1}. Methods for BN structure learning can be divided into score-based learning and constraint-based learning. Because constraint-based learning faces two assumptions that cannot be satisfied easily in nature, score-based learning is considered as a more important and stable way for BN structure learning. But the search process of these methods is proved to be an NP-hard problem \cite{expert,BN-AIIM,EX3}, that can be solved by combination optimization. Although researchers have proposed several heuristic frameworks for the search process recently, the score-based BN structure learning still faces major difficulties \cite{BN-001,BN-001,BN-003}. 

Heuristic algorithms are widely considered to solve many COPs, including BN structure learning. These heuristic algorithms can be generally divided into four categories \cite{CHIO}: evaluation-based (genetic algorithm \cite{bib9}, particle swarm optimization \cite{bib10}), swarm-based (ant lion optimizer \cite{bib21}, whale optimization algorithm \cite{bib7}), physical-based (simulated annealing \cite{bib13}, sine cosine algorithm \cite{bib36}, max min hill climbing \cite{MMHC}) and human-based algorithms (fireworks algorithm \cite{bib15}, harmony search \cite{bib14}). A difficult problem faced by these algorithms is that the parameters of the algorithms are difficult to select. For example, for simulated annealing algorithms, researchers need to consider the parameters such as annealing coefficient, warming coefficient, warming times and maximum iteration times. Any improper parameter will affect the accuracy of the final built BN. The second problem is that these algorithms are seriously homogeneous. For example, most swarm-based algorithms generate several individual solutions. And in the process of their iteration, these solutions jointly move towards greater fitness. Researchers have created different algorithms based on the living habits of different animals. Although these algorithms go through different iterative processes in the whole iterative process, there is not much difference on the whole, and researchers will fall into the dilemma of wondering which heuristic algorithm to choose. 

Methods based on reinforcement learning (RL) have been widely applied to solve COPs recently. Grinsztajn {\it et al.} proposed Poppy \cite{RL05}, a simple training procedure for populations, that induces an unsupervised specialization targeted solely at maximizing the performance of the population. They showed Poppy produces a set of complementary policies and obtains state-of-the-art results on three popular NP-hard problems: traveling salesman, capacitated vehicle routing, and job-shop scheduling. Wang {\it et al.} \cite{RL03} proposed utilizing a single deep reinforcement learning model to solve COPs, using the well-known multi-objective traveling salesman problem (MOTSP) as an example. Their method employs an encoder-decoder framework to learn the mapping from the MOTSP instance to its Pareto-optimal set. Additionally, they developed a top-k baseline to enable more efficient data utilization and lightweight training. By comparing their method with heuristic-based and learning-based ones on MOTSP instances, the experimental results show that their method can solve MOTSP instances in real-time and outperform the other algorithms, especially on large-scale problem instances. Achamrah \cite{RL01} introduced a novel framework that combines the adaptability and learning capabilities of deep reinforcement learning (DRL) with the efficiency of transfer learning and neural architecture search. This framework enables the leveraging of knowledge gained from solving COPs to enhance the solving of different but related COPs, thereby eliminating the necessity for retraining models from scratch for each new problem variant to be solved. These show that applying RL to COPs is more than attractive since it removes the need for expert knowledge or pre-solved instances and reduces the occupation of people. However, to the best of our knowledge, there is no method based on reinforcement learning to solve the searching for BN structure learning, nor method that consists of the idea of reinforcement learning. To this end, the work attempts to guide the search process of BN structure learning from the perspective of reinforcement learning. 

In this paper, a novel score-based \textbf{Bayes}ian network structure learning algorithm based on \textbf{R}einforcement \textbf{L}earning search strategy (\textbf{RLBayes}) is proposed, and more specifically, based on $Q$-learning. Because the search space of Bayesian network is infinite, it is unrealistic to create and maintain $Q$-table if the general $Q$-learning framework is adopted. To overcome this problem, a dynamic $Q$-table maintaining method is proposed, which tends to use the limited space to represent the infinite search space. The iteration process is recorded and guided by this dynamically maintained $Q$-table, which can converge to the optimal solution and avoid local optimal solutions. RLBayes is theoretically proved to be able to converge to the global optimal solution when the parameters are reasonably set. And experiments on numerous benchmark datasets prove that RLBayes shows better learning performance than other heuristic algorithms. 

\textbf{Overall, the advantages of RLBayes can be summarized as: }

\begin{itemize}
\item RLBayes is a BN structure learning algorithm via a RL-inspired search strategy, that tries to search the BN close enough to the global optimal solution within a huge search space infeasible to traverse completely.
\item The BN searched by RLBayes is much better on the performance of F1 score and AUC, compared to the BNs searched by other heuristic methods; 
\item The parameter of RLBayes is easy to set, the increase of either the maximum of the iterations or the maximum size of $Q$-table will make the results better. So it is possible to pursue better results by enlarging them, if given enough computation time and RAM memory.
\end{itemize}

The rest of the paper is organized as: In \autoref{section2}, the structure learning of BN is discussed. In \autoref{section3}, RLBayes for the structure learning of BNs is introduced. In \autoref{section4}, experiments are conducted, which comfirms the advantages of RLBayes. Last in \autoref{section5}, the conclusions are given. 

\section{Preliminary}
\label{section2}

Bayesian Networks (BNs) have recently gained increasing popularity as a versatile tool for addressing uncertainty in a wide array of fields, including medicine, economics and others~\cite{ZR,ICASSP3}. Their utility is particularly evident in real-world scenarios where intricate questions necessitate hypothetical evidence to guide intervention strategies \cite{SIGA-BN}. Nevertheless, learning the graphical structure of a BN poses a significant challenge, particularly when formulating causal models for complex problems. In this section, the two categories for the structure learning of BNs is reviewed, constraint-based methods and score-based methods. Specifically, we concentrate on the score-based structure learning methods, because we make contributions mainly for this kind of methods in this paper. 

\subsection{Constraint-Based Structure Learning of BN}

Constraint-based BN Structure Learning uses conditional independence (CI) tests on the dataset to determine whether two selected variables are related according to the given dataset, and hence learn a graph consistent with the dataset. The CI tests rely heavily on statistical or information measures and the learning results depend on the threshold selected for CI tests \cite{constraint}. For instance, the recursive autonomy identification (RAI) algorithm achieves high-order CI tests using sequentialization and recursion, thereby learning the structure of BN \cite{RAI}.  However, constraint-based methods are built upon two assumptions: the directed CI model should be faithful to a directed acyclic graph (DAG), and the directed CI tests performed on data should accurately reflect the independence model, which are hard to be satisfied in reality \cite{unstable}. So constraint-based methods are unstable and prone to cascade effects where an early error in the learning process will bring about a quite different DAG structure. \cite{twoassumtion}.

\subsection{Score-Based Structure Learning of BN}
Score-based structure learning is the other main class of BN structure learning and consists of two elements: (a) an objective function that can be used to evaluate each network explored in the search space of networks; (b) a search strategy that determines which path to follow in the search space of possible networks. The objective function is the optimization target function. The higher the score function is, the better the model fits the dataset. But the search process of score-based structure learning is proved to be NP-hard, making it a COP. And it's infeasible to search all possible networks when the nodes of the network are more than six. One of the overriding challenges for score-based learning is to find high, or ideally the highest, scored networks among the vast number of possible networks \cite{CPD-NSL}. 

The objective function of score-based methods is the objection to be optimized. A branch of score functions are based on penalized log-likelihood (LL) functions. Under the standard independent and identically distributed assumption, the likelihood of the dataset $\{D_1,...,D_N\}$ given a structure $B$ can be calculated as: 
\begin{equation}\label{align02}
LL(D|B)=\sum\limits_{j=1}^{N}{\log ({{D}_{j}}|B)}
\end{equation}
Adding edges will usually increases the log-likelihood (\autoref{align02}), leading to at least two troubles. First, it may lead to an overfitting model and result in poor performance. Second, densely connected networks will increase the running time. Adding a penalty on log-likelihood (\autoref{align02}) aims to address these problems by adding a penalty term which prevents complex networks. This has led to a class of score functions that share the first item (log-likelihood term) but differ only in the second (penalty term). Suppose $k$ represents the number of model parameters and $n$ represents the number of samples, Akaike information criterion (AIC) and Bayesian information criterion (BIC) is defined as: 
\begin{align}
&AIC = LL(D|B) - 2 \times k \\
&BIC = LL(D|B) - ln(n) \times k \label{align03}
\end{align}
The penalty term of BIC is larger than that of AIC, and the number of samples is taken into account. When the number of samples is too large, the model complexity caused by the excessive accuracy of the model can be effectively prevented. BIC tends to favor sparse structures than AIC. Other types of objection functions, such as BDe and MDL can be referred to \cite{BNL}.

Meanwhile, for score-based methods, the searching process is usually formulated as a COP, because the search process is an NP-hard problem. The search strategy is usually guided by various heuristics, very convenient hill-climbing, but many other heuristics including  particle swarm optimization, genetic algorithm have also been usually considered. Take genetic algorithm as an example, each individual in the population represents a BN and the objection function to be optimized is one of the Bayesian score functions mentioned above. At each iteration, the well-scored BNs are selected to cross and generate new individuals, and sometime mutations happen on some individuals. These mutations include adding, deleting and reversing an edge in an individual. After iterating for several times, the best scored individual in the population is chosen as the optimization result. Despite the success of these methods in searching for the best-fit BN, the constructing accuracy of the BN is quite limited. And the parameters of these heuristics methods is usually vast and hard to choose. As a result, the need of accurately building BNs is always calling for brand new heuristic search strategy for BN structure learning \cite{EX1}.  

\subsection{Operations and Reverse Operations}\label{ssec}
\begin{figure}[!t]
	\centering  
		\includegraphics[width=0.9\columnwidth]{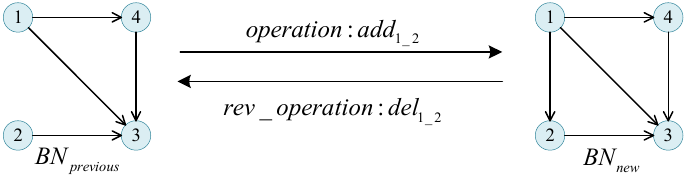}
\caption{Operation and Reverse Operation}\label{rev}
\end{figure}

The operation is a common concept in the search process of score-based BN structure learning. For the convenience of the following description, we first introduce the operation and reverse operation considered in this paper. There are three operations involved in this paper, including adding edges, deleting edges and reversing edges. For example, $del_{1\_2}$ means deleting the edge 1 $\rightarrow$ 2. Reverse operation refers to the operation that can be performed to return from the resulting network by the operation to the previous network. Taking \autoref{rev} as an example, the reverse operation of executing $add_{1\_2}$ on $BN_{previous}$ to get $BN_{new}$ is to execute $del_{1\_2}$ on $BN_{new}$. It is a theorem that for any operation, its reverse operation always exists. Moreover, if an operation can be conducted, its corresponding reverse operation can be conducted as well. 

\section{RLBayes}\label{section3}
In this section, the proposed BLBayes method is fully discussed. Then, the global convergence of RLBayes is proved. Lastly, the implementation of RLBayes is discussed. 

\subsection{Overall Procedure of RLBayes}  

\begin{figure*}[!t]
	\centering  
		\includegraphics[width=0.9\textwidth]{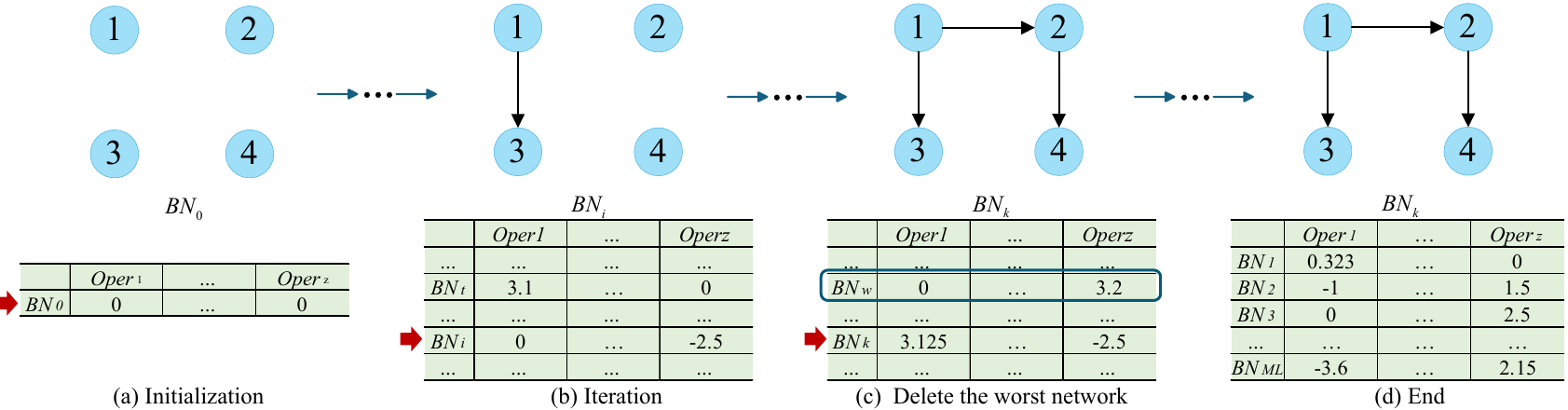}
\caption{Overall iteration procedure of RLBayes}\label{frame}
\end{figure*}

The procedure of RLBayes starts from an empty Bayesian network, as is presented in \autoref{frame}(a), and keeps maintaining a dynamic $Q$-table, denoted as $table_q$. For $table_q$, each row index is a BN, while each column represents a possible operation on the corresponding BN of the row. Each value in the corresponding position in $table_q$ represents the benefit that the operation would bring if performed on the BN of the row. For a BN, sometimes an operation cannot be conducted on it because of the acycle restriction or the existence/non-existence of an edge. RLBayes requires that the possible operations for any BN is the same, no matter these operations can be conducted on the BN or not. In this way, any position in $table_q$ is not $null$ and each row of $table_q$ is of the same length. Then the corresponding value of an unenforceable operation of a BN will be updated to -$inf$, after failing to conduct the operation. This means the conduction of the operation will result in benefit of `-$inf$'. And in this way, the operations with benefit `-$inf$' will be never chosen again.

Then consider how to choose an operation at each iteration. Suppose the iteration comes to the $ind_a$-th row, and the BN of this row is denoted as $BN_p$. The $ind_a$-th row of $table_q$ is exacted. Then the are two possible cases. For the first case, the operation is chosen according to the benefit the operations will bring. The operations with higher benefits are more likely to be chosen, the operations with lower benefits are less likely to be chosen and the operations with `-$inf$' benefits are never chosen. For the other case, an operation is chosen randomly from the operations with value 0. This is to help choose an operation at the beginning of the procedure or to avoid being trapped into a local optimal solution. The operation choosing process is detailed in Algorithm \autoref{alg0}.

\begin{algorithm}[t]
    \caption{Operation Choosing}\label{al01}
    \textbf{Input}: the $Q$-table $table_q$, the current iteration index $ind_a$\\
    \textbf{Output}: $oper$;
    \begin{algorithmic}[1] 
			\label{alg0}
			\IF {rand\_uniform() $\geq$ 0.5} 
			\STATE $oper$ $\leftarrow$ an random operation among all the possible operations of the BN in the $ind_a$-th row of $table_q$. 
			\ELSE
			\STATE $oper$ $\leftarrow$ an operation according to the benefit value of BN in the $ind_a$-th row of $table_q$.
			\ENDIF
			\STATE \textbf{return} $oper$.
    \end{algorithmic}
\end{algorithm} 

Once a specific operation $oper$ is decided for $BN_p$, there are two possible upcoming cases then. The first case is the $oper$ can be conducted, yielding a new BN different from $BN_p$, which is denoted as $BN_n$. Then there are also two possible scenarios. The first scenario is the newly generated $BN_n$ has already been in the row index of $table_q$. Suppose the index of $BN_n$ is $ind_b$. Then the thing to do is to modify the related value in $Q$-table $table_q$ as: 
\begin{align}\label{a3}
&table_q\left[ind_a, oper\right] \\ \nonumber
=&Score\left(BN_n, D\right) - Score\left(BN_p, D\right)
\end{align}
\begin{align}\label{a4}
&table_q\left[ind_b, rev\_oper\right] \\ \nonumber
=&Score\left(BN_p, D\right) - Score\left(BN_n, D\right)
\end{align}
where $Score\left(B, D\right)$ means the Bayesian score function of BN structure $B$ given the dataset $D$, $rev\_oper$ is the reverse operation on $BN_n$ of $oper$ on $BN_p$. According to the discussion in \autoref{ssec}, $rev\_oper$ exists and can be performed on $BN_n$. And it is obvious that the benefits of $oper$ on $BN_p$ and the one of $rev\_oper$ on $BN_p$ are opposite each other. The other scenario is that the newly generated BN has not already been in the row index. Then it is to add a new row in the $Q$-table $table_q$, initialize the benefit value as zero and modify the related values as: 
\begin{align}\label{a1}
&table_q\left[ind_a, oper\right] \\ \nonumber
=&Score\left(BN_n, D\right) - Score\left(BN_p, D\right)
\end{align}
\begin{align}\label{a2}
&table_q\left[len(table_q) - 1, rev\_oper\right] \\ \nonumber
=&Score\left(BN_p, D\right) - Score\left(BN_n, D\right)
\end{align}

The other case is the operation can't be conducted at all because of the acyclic restriction of BNs or others. For this case, only setting the operation value of BN -$inf$ is needed, in which way the operation will not been chosen next time when iteration comes to $BN_p$. 
\begin{equation}table_q\left[ind_a, oper\right] = -inf\end{equation}

An example of the above process is clearly presented in Example \autoref{ex1}. Besides, once a BN is generated, the new operation is chosen according to the benefit value of the newly generated BN and conducted on this newly generated BN. As the process repeats continually, more and more new BNs are generated and the length of $table_q$ grows larger and larger. However, it is impossible to store all of the generated BNs all the time, since the structure learning of BN is NP-hard. And if the length of $table_q$ is too large, the process of detecting whether the newly generated BN already exists in $table_q$ is too time-consuming because it is needed to traverse all BNs in $table_q$. As the iteration goes on, it is considered that the newly generated BN is close to the BN it is generated from with a pretty good score, rather than a badly scored one. So, RLBayes adopts a greedy strategy to limit the length of $table_q$, by setting a maximum length of $table_q$ as $MAX\_LENGTH$. Once the length of $table_q$ reaches $MAX\_LENGTH$, the row with the worst scored BN is dropped. (In \autoref{frame}(c), the length of $table_q$ reaches $MAX\_LENGTH$ and the row with the worst scored BN is deleted, boxed in blue.) After enough iterations, the best-scored BN in $table_q$ is regarded as the search result.

\begin{figure}[t]
	\centering  
	\subfigbottomskip=-1pt 
	\subfigcapskip=0pt 
	\subfigure[Operations on BN]{\label{figure2-1}
		\includegraphics[width=0.95\columnwidth]{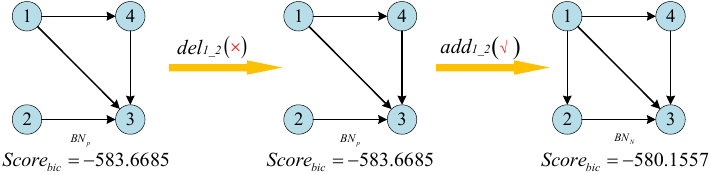}}
	  \\
	\subfigure[The maintenance of $table_q$ along with the operations]{\label{figure2-2}
		\includegraphics[width=0.95\columnwidth]{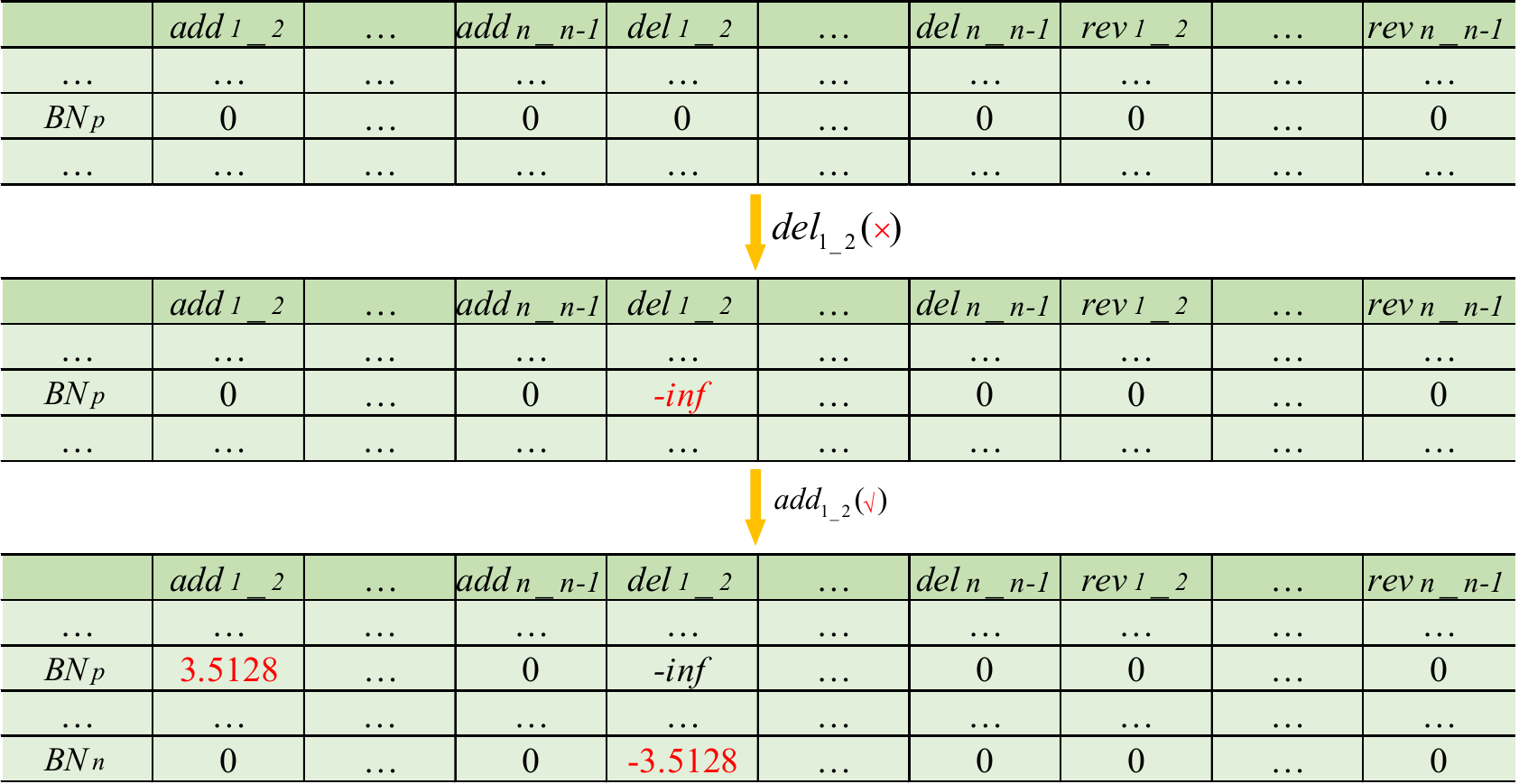}}
\vspace{4pt}
\caption{An example of the maintenance of $table_q$}\label{figure2}
\end{figure}

\begin{example}
\label{ex1}
As is depicted in \autoref{figure2}, the BN on the far left in Subfigure \autoref{figure2-1} is denoted as $BN_p$, and its state value is shown in the table on the top in Subfigure \autoref{figure2-2}. Then an operation is chosen according to Algorithm \autoref{al01}. Suppose the operation `$del_{1\_2}$' is chosen here. Then this operation cannot be conducted because there is no edge linking from 1 to 2. Then it is needed to update the `$del_{1\_2}$' of $BN_p$ to `-$inf$', so that when the iteration comes to $BN_p$ again, this operation cannot be chosen. Then the next iteration continues on $BN_p$ and suppose the operation `$add_{1\_2}$' is chosen this time. This operation can be conducted because it vialote none of the restrictions of BNs. Then the newly generated BN is denoted as $BN_n$, which scored -583.6685. Compared to $BN_p$, $BN_n$ gains the benefit of 3.5128. So the benefit of `$add_{1\_2}$' on $BN_p$ is updated to 3.5128, which means adding the edge 1$\rightarrow$2 will make 3.5128 benefits on the score function. Meanwhile, the benefit of `$del_{1\_2}$' on $BN_n$ is set as -3.5128, and other benefit values of $BN_n$ are initialized as zero. Then the iteration continues on $BN_n$ and the operation of `$del_{1\_2}$' will be less likely chosen because it will cause a negative benefit. 
\end{example}

\subsection{Proof of Global Convergence of RLBayes}\label{subsecproof}
As a heuristic optimization method, RLBayes can achieve global optimization Bayesian network, if it meets the following two conditions~\cite{FTTA}:
\begin{itemize}
    \item RLBayes has the ability to achieve local optimization; 
    \item RLBayes has the ability to transfer from one local optimal state to another better local optimal state. 
\end{itemize}
\vspace{-10pt}
\begin{proof}
	When RLBayes constructs a new solution (BN), the new solution will be saved as a new row in the $Q$-table no matter the score function of the new solution is better or not. While recording the new state, the benefit of the past operation (a positive/negative real value or -$inf$) will be calculated and recorded in $table_q$. Then the same procedure repeats for a great number of times. As the length of the $Q$-table exceeds the maximum limit, the row in $table_q$ with the worst scored BN is deleted. At each iteration, the operations with higher benefits are more likely to be chosen to generate new solutions than the ones with lower benefits and the operations with -$inf$ benefits are never chosen. By doing so, RLBayes can converge to the local optimization as the iterations go on. \par
For one local optimal solution, almost each operation of it is negative or -$inf$. In such case, RLBayes won't stop but keeps seeking for a better state. Even though the newly generated solution is worse than the local optimal solution, the new solution will be stored and recorded. Then many of the benefits of operations of the new solution is calculated. When the length of $table_q$ is larger than $MAX\_LENGTH$. The row in $table_q$ with the worst Bayesian score function will be eliminated, while the local optimal solution along with other good solutions will be maintained. Once a better state is found, the BNs in the $Q$-table tends to get closer to the better state.  Therefore, RLBayes can transfer from one local optimal state to another better local optimal state. 
\end{proof}

\subsection{Implementation of RLBayes}

\begin{algorithm}[!t]
    \caption{Implementation of RLBayes}
    \textbf{Input}: Dataset to be inferred from $D$, maximum iteration times $MAX\_ITER$, maximum $Q$-table length $MAX\_LENGTH$, the state transferfactor $\theta$;\\
    \textbf{Output}: a learned BN $BN_{result}$;
    \begin{algorithmic}[1] 
			\label{alg2}
			\STATE Initialize an empty BN $BN_0$, $i$=0, $cur\_index$=0;
			\STATE Initailize $Q$-table $table_q$;
			\FOR{$i$ in $1 : MAX\_ITER$}
			\STATE $BN_p$ $\leftarrow$ the BN in the $cur\_index$-th row of $table_q$;
			\STATE $oper$ $\leftarrow$ choose($table_q$, $cur\_index$);
			\IF {$oper$ is conducted}
			\STATE The newly generated BN is denoted as $BN_n$;
			\IF {$BN_n$ does not exist in $table_q$}
			\STATE add a new row for $BN_n$;
			\STATE $table_q$[$cur\_index$, $oper$] = score($D$, $BN_n$) - score($D$, $BN_p$);
			\STATE $table_q$[$len$($table_q$) - 1, $rev\_oper$] = score($D$, $BN_p$)-score($D$, $BN_n$);
			\STATE $cur\_index$ $\leftarrow$ $len$($table_q$) - 1
			\ELSE
			\STATE $BN_n\_index$ = index($table_q$, $BN_n$)
			\STATE $table_q$[$cur\_index$, $oper$] = score($D$, $BN_n$) - score($D$, $BN_p$);
			\STATE $table_q$[$BN_n\_index$, $rev\_oper$] = score($D$, $BN_p$)-score($D$, $BN_n$);
			\STATE $cur\_index$ $\leftarrow$ $BN_n\_index$
			\ENDIF
			\ELSE
			\STATE $table_q$[$cur\_index$, $oper$] = -$inf$;
			\ENDIF
			\WHILE{len($table_q$) $\geq$ $MAX\_LENGTH$}
			\STATE Eliminate the row in $table_q$, whose corresponding $BN$ scores the worst;
			\ENDWHILE
			\IF {$rand\_uniform(0, 1)$ $\leq$ $\theta$}
			\STATE $cur\_index$ $\leftarrow$ the index of the best-scored BN in $table_q$;
			\ENDIF
			\ENDFOR
			\STATE $BN_{result}$ $\leftarrow$ the best-scored BN in $table_q$;
			\STATE \textbf{return} $BN_{result}$.
    \end{algorithmic}
\end{algorithm}

\begin{table}[!t]
\renewcommand{\arraystretch}{0.97}
    \centering
    \caption{Details of Networks}
	\vspace{-6pt}
    \begin{tabular}{lccc}
        \toprule
        Network  & No. nodes & No. edges & Density \\
        \midrule
        asia      & 8         & 8       &14.28\%    \\
        sachs     & 11        & 17      &15.45\%    \\
        child     & 20        & 25      &6.579\%    \\
        insurance & 27        & 52      &7.407\%    \\
        hailfinder& 56        & 66      &2.143\%    \\
		win95pts  & 76        & 112     &1.965\%    \\
        \bottomrule
    \end{tabular}
    \label{tab:booktabs}
\end{table}

\begin{table*}[!t]
\centering
\renewcommand{\arraystretch}{0.98}
\caption{F1-Score Comparison of the construction BNs via different heuristic methods}\label{table2}
\vspace{-6pt}
\begin{tabular}{ccccccc}
\toprule
                 & asia                       & sachs                      & child                      & insurance                           & hailfinder                 & win95pts               \\
\midrule
GA               & 0.3462$\pm$0.1438          & {\ul 0.5932$\pm$0.1230} & 0.5882$\pm$0.1289          & 0.3279$\pm$0.0377 &0.0602$\pm$0.0328              & 0.0525$\pm$0.0142         \\
HC               & 0.5350$\pm$0.2094          & \textbf{0.6095$\pm$0.0714}          & {0.6333$\pm$0.0167}        &  0.3328$\pm$0.0085                &  \textbf{0.3285$\pm$0.0098}  &  {\ul 0.2768$\pm$0.0361}                      \\
SA               & 0.2622$\pm$0.1579          & 0.5017$\pm$0.1550          & 0.5203$\pm$0.1137          & {\ul0.3455$\pm$0.0582}          & 0.2946$\pm$0.0566          & 0.1984$\pm$0.0221      \\
IWO              & 0.4135$\pm$0.2506          & 0.5410$\pm$0.0780    & 0.5538$\pm$0.1217          & 0.3176$\pm$0.0893                   & {\ul 0.3282$\pm$0.0591}          & 0.2120$\pm$0.0326      \\
FTTA             & 0.5142$\pm$0.2523          & 0.3518$\pm$0.1370          & 0.2543$\pm$0.0492          & 0.1374$\pm$0.0405                   & 0.0304$\pm$0.0206    & 0.0484$\pm$0.0160            \\
BNC-PSO          & {\ul 0.5434$\pm$0.1598}    & 0.4589$\pm$0.1021          & \textbf{0.6938$\pm$0.0609} & 0.2209$\pm$0.0501                   & 0.2184$\pm$0.0421    & 0.2570$\pm$0.0447          \\
\textbf{RLBayes} & \textbf{0.6283$\pm$0.1950} & 0.5851$\pm$0.0659          & {\ul 0.6863$\pm$0.0980}    & \textbf{0.3939$\pm$0.0688} & 0.3188$\pm$0.0467 & \textbf{0.2838$\pm$0.0440}         \\
\bottomrule
\end{tabular}
\end{table*}

\begin{table*}[!t]
\centering
\renewcommand{\arraystretch}{0.98}
\caption{AUC Comparison of the construction BNs via different heuristic methods}\label{table3}
\vspace{-6pt}
\begin{tabular}{ccccccc}
\toprule
                 & asia                       & sachs                      & child                      & insurance                   & hailfinder                 & win95pts                   \\
\midrule
GA               & 0.6250$\pm$0.0814          & {\ul 0.7510$\pm$0.0669}    & 0.7749$\pm$0.0626          & 0.6237$\pm$0.0171	         &  0.5198$\pm$0.0168	       & 0.5176$\pm$0.0079          \\
HC               & 0.7259$\pm$0.1120          & \textbf{0.7553$\pm$0.0391} & 0.7936$\pm$0.0085          &   0.6295$\pm$0.0095         &  0.6470$\pm$0.0047         &0.6308$\pm$0.0178           \\
SA               & 0.5768$\pm$0.0897          & 0.7011$\pm$0.0843          & 0.7421$\pm$0.0564          & {\ul 0.6308$\pm$0.0265}     & 0.6367$\pm$0.0251          & 0.6260$\pm$0.0162          \\
IWO              & 0.6589$\pm$0.1394          & 0.7216$\pm$0.0426          & 0.7561$\pm$0.0598          & 0.6188$\pm$0.0410           & \textbf{0.6511$\pm$0.0288} & {\ul0.6348$\pm$0.0229}      \\
FTTA             &  0.7134$\pm$0.1394         & 0.6195$\pm$0.0740          & 0.6112$\pm$0.0284          & 0.5358$\pm$0.0209          & 0.5049$\pm$0.0106           &  0.5148$\pm$0.0078          \\
BNC-PSO          & {\ul 0.7411$\pm$0.0914}    & 0.6776$\pm$0.0548          & {\ul 0.8219$\pm$0.0292}    & 0.5734$\pm$0.0220          & 0.5982$\pm$0.0208           &        0.6250$\pm$0.0226    \\
\textbf{RLBayes} & \textbf{0.7768$\pm$0.1034} & 0.7436$\pm$0.0353          & \textbf{0.8228$\pm$0.0460} & \textbf{0.6527$\pm$0.0315} & {\ul 0.6477$\pm$0.0216}     & \textbf{0.6423$\pm$0.0235}  \\
\bottomrule
\end{tabular}
\end{table*}
\vspace{-3pt}
The implementation of RLBayes is shown in Algorithm \autoref{alg2}. The input contains the dataset to be inferred from $D$, maximum iteration times $MAX\_ITER$, maximum $Q$-table length $MAX\_LENGTH$ and the state transfer factor $\theta$. The output contains only a learned BN $BN_{result}$. The algorithm starts with an empty Bayesian network $BN_0$, $i$ is a variable that counts the number of iterations, $cur\_index$ records the index of the row which current iteration comes to. So the initial value of $i$ and $cur\_index$ is zero (line 1). Then initialize the $Q$-table $table_q$, whose length is one at this time. The only BN in $table_q$ is $BN_0$ at row index zero currently. And each value at the row is initialized as zero (line 2). In each iteration, the BN in the row $cur\_index$ is denoted as $BN_p$. Then choose an operation $oper$ for $BN_p$ according to the values in $table_q$[$cur\_index$] and Algorithm \ref{alg0} (lines 4-5). If $oper$ can be conducted, the newly generated BN is denoted as $BN_n$. If $BN_n$ does not exist in $table_q$, update the values in the rows $cur\_index$ according to Equation \ref{a1} and append a new row for $BN_n$. All values of the new row is initialized as zero, and update the corresponding value according to Equation \ref{a2} (lines 8-12). If $BN_n$ exists in $table_q$, then update the values in the rows $cur\_index$ and $BN_n\_index$ according to Equations (\ref{a3}-\ref{a4}), where $BN_n\_{index}$ is the index of $BN_n$ in $table_q$ currently (lines 14-17). If $oper$ cannot be conducted, only a corresponding value should be set -$inf$ (line 20). Then if the length of $table_q$ is larger than $MAX\_LENGTH$, eliminate the row in $table_q$, whose corresponding $BN$ scores the worst repeatedly until the length of $table_q$ is less than $MAX\_LENGTH$ (lines 22-24). At last, to make the BNs in $table_q$ closer to the global optimal solution quickly, $cur\_{index}$ is transferred to the index of the best-scored BN in $table_q$ with the possibility of $\theta$ (lines 25-27). After the whole iterations, the best-scored BN in $table_q$, denoted as $BN_{result}$, is returned as the search result.
\vspace{-4pt}
\section{Experiments}\label{section4}
The experiments mainly focus on answering three research questions (RQs):

\textbf{(RQ. 1)} Does RLBayes show better ability to seek for the best-fit BN, compare to other heuristic methods within the same, or nearly the same time consumption?

\textbf{(RQ. 2)} Do the parameters of RLBayes easy to control, and does the effect of RLBayes easily affected by the change of these parameters? 

\textbf{(RQ. 3)} How does the network construction accuracy change with the change of network scale or density? 

To answer the research questions, the benchmark models chosen are listed in \autoref{tab:booktabs}. Among them, $asia$ and $sachs$ are small networks, $child$ and $insurance$ are networks of medium size and  $hailfinder$, and $win95pts$ are examples of large networks. Samples are generated according to the networks, which are used to score and search for the best-fit network. Baseline heuristic methods, including hill climbing (HC), gene algorithm (GA), invasive weed optimization (IWO) \cite{IWA}, football team training algorithm (FTTA) \cite{FTTA}, simulated annealing (SA)\cite{bib13}, Bayesian network construction algorithm using PSO (BNC-PSO) \cite{BNC-PSO} are chosen for comparison with RLBayes. To maintain fairness, the parameters of the baselines are set either their default settings (if provided) or the proper parameters that make the runtime quite close to the runtime of RLBayes. And the BIC score function is selected as the objection function for optimization. The experiments are conducted on a PC with an Inter(R) Core(TM) i5-14600KF CPU and 64 GB RAM. The operation system is Windows. 
\vspace{-2pt}
\subsection{Network Construction Accuracy Comparison}
To eliminate the impact of individual successes or failures on the conclusions of the experiments, each search method is performed on each dataset ten times, and the mean value and variance value of the evaluation metrics for the search results are calculated for comparison. In \autoref{table2} (F1) and \autoref{table3} (AUC), the best performance among these methods is presented in bold and the second-best performance among these methods is presented underlined. First, RLBayes gains an average ranking of 1.83 and 1.5 on F1 and AUC, respectively, which means that it achieves nearly the best performance among the considered heuristic search methods. And RLBayes has never been out of the top three among the methods. Second, though  some evolution-based algorithms and swarm-based algorithms achieve better performance than RLBayes on some of the networks such as $sachs$, the performance of these methods experiences a sharp decrease, with the increase of the number of nodes. For the networks with more than 50 nodes, the AUC of these methods drop even nearly to 0.5. On the opposite, RLBayes is more stable with the increase of the number of nodes. Overall, RLBayes shows better ability to seek for the best-fit BN, compare to other methods in the same, or nearly the same time consumption.
\vspace{-2pt}
\subsection{Analysis of the Parameters of RLBayes}
\begin{figure}[!t]
	\centering  
	\subfigbottomskip=-8pt 
	\subfigcapskip=-5.5pt 
	\subfigure[asia]{\label{figure4-1}
		\includegraphics[width=0.9\columnwidth]{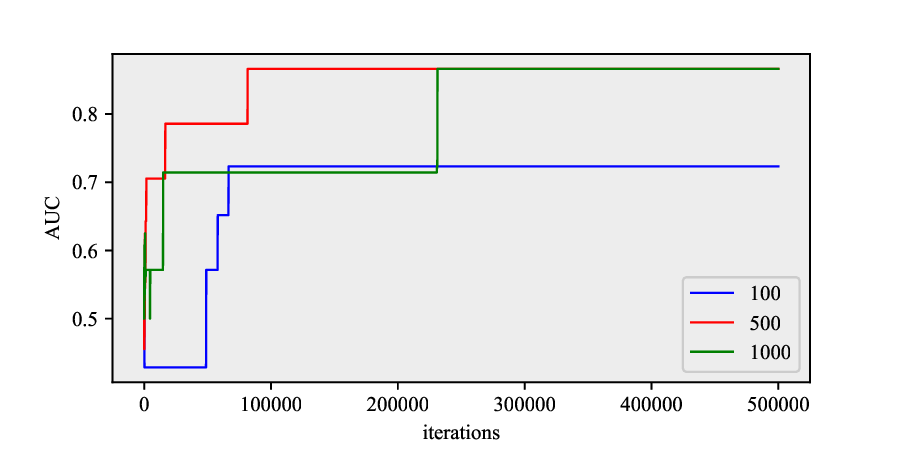}}
	\subfigure[insurance]{\label{figure4-2}
		\includegraphics[width=0.9\columnwidth]{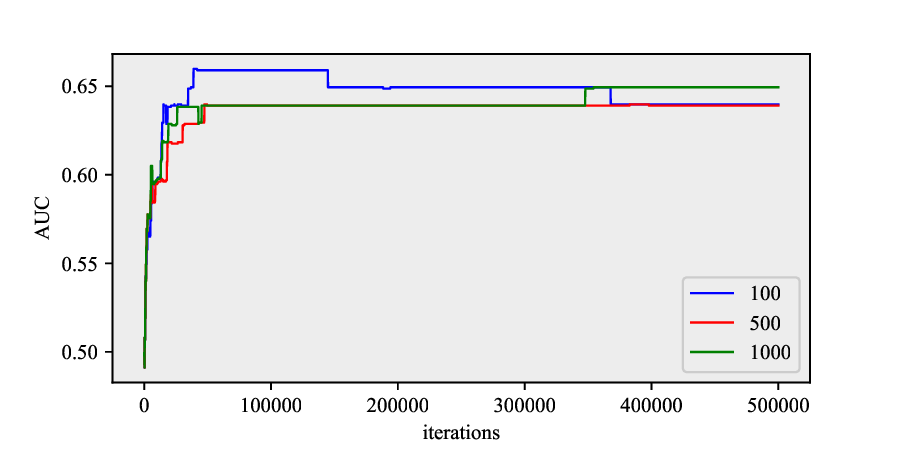}}
	\subfigure[hailfinder]{\label{figure4-3}
		\includegraphics[width=0.9\columnwidth]{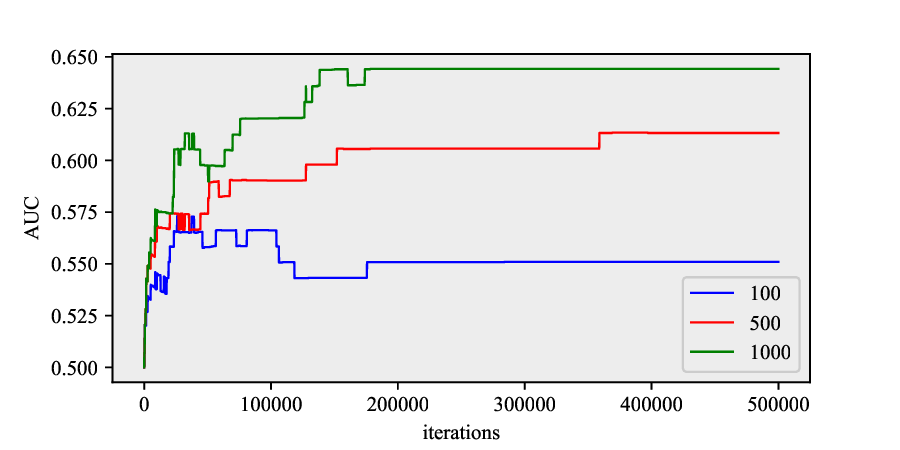}}
\vspace{-2pt}
\caption{Analysis of the Parameters of RLBayes}\label{figure-result}
\end{figure}

As is mentioned in \autoref{sec1}, the parameters of heuristic methods are hard to select. In the face of the complex BN structure learning, the parameter selection of the heuristic methods has more influences on the final results. RLBayes has two major control parameters, $MAX\_LENGTH$ and $MAX\_ITER$. Herein, we make the analysis of how much the parameters of RLBayes will affect the final BN structure learning results. As shown in \autoref{figure-result}, the experiments are carried out independently on three networks. The $x$-axis of each subplot represents the iterations ranging from 0 to 500000, while the $y$-axis of each subplot is the AUC value of the constructed network. The legends `100', `500' and `1000' represents different $MAX\_LENGTH$ setting. 

First, for three networks, the AUC value of the constructed network generally increases with the number of iterations. After several iterations, it is possible to fall into a local optimal solution. And as the iteration continues, it is also possible to jump out of the local optimal solution to better one. This result agrees with the convergence proof in \autoref{subsecproof}. Secondly, it is shown that the AUC value of the constructed network generally increases with the $MAX\_LENGTH$. This is because as the $MAX\_LENGTH$ increases, the ability of storing BNs is better. The traversed BNs so far need not to be deleted given a larger $MAX\_LENGTH$, but are stored to guide the future `$operation$'s. Therefore, a larger length leads to more accurate results. Overall, the increase of either $MAX\_LENGTH$ or $MAX\_ITER$ will lead to better performance of RLBayes, thereby making controlling the parameters easier than many other heuristic methods.

\subsection{Network Construction Accuracy Change with the Change of Network Sparsity}
	As is pointed out by \cite{BICLP,Guo}, the network construction accuracy decreases with the increase of network density via different models. Our experiments use BIC score function and different heuristic methods to learn BN. The results have the same trend as those of \cite{BICLP,BICLasso}. As shown in \autoref{tab:booktabs}, the sparsity of the network increases with increase of the number of nodes As shown in \autoref{table2}-\ref{table3}, no matter which heuristic algorithm used to search the  BN, the F1 score and AUC obtained decrease as the sparsity of the network increases. However, compared to the baseline  heuristic methods, RLBayes is more stable as it decreases much slower than baselines, making it more attractive and competitive. These results agree with the previous conclusions and confirms the correctness of our coding and the soundness of parameter selection. 

\section{Conclusion}\label{section5}
This paper proposes RLBayes, a BN structure learning algorithm via RL-inspired search strategy, that tries to search the BN close enough to the global optimal solution within a huge search space infeasible to traverse completely. In order to solve the problem that the solution space in the structure learning process of BNs based on score functions is close to infinite, a dynamic $Q$-table is created and maintained to store a certain amount of possible BNs and the benefit values of possible operations on these BNs in limited space. Theoretically, it is strictly proved that RLBayes has the ability to converge to the global optimal solution when the parameters are set reasonably. Experimentally, experiments on multiple complex benchmark datasets of different scales confirm that the BNs constructed by RLBayes are more accurate than those constructed by other heuristics. It is also proved that the parameters of RLBayes are easy to control. In conclusion, RLBayes is an attractive score-based BN structural learning algorithm based on reinforcement learning. 
\section*{Data \& Codes Available}
Data and codes are available at appendix, and will be open source after acceptance. 
\bibliographystyle{named}
\bibliography{ijcai25}
\end{document}